\documentclass[runningheads]{llncs}
\usepackage{eccv}
\usepackage{eccvabbrv}
\usepackage{graphicx}
\usepackage{booktabs}
\usepackage[accsupp]{axessibility}  
\usepackage{hyperref}
\usepackage{orcidlink}

\begin{document}

\title{DALDA: Data Augmentation Leveraging Diffusion Model and LLM with Adaptive Guidance Scaling}

\titlerunning{Data Augmentation Leveraging DM and LLM Prompts}

\author{Kyuheon Jung\inst{1}\orcidlink{0000-0002-5657-7533} \and
Yongdeuk Seo\inst{1}\orcidlink{0009-0003-3399-8364}  \and
Seongwoo Cho\inst{1}\orcidlink{0009-0005-5942-5358}  \and
Jaeyoung Kim \inst{2}\orcidlink{0000-0003-0880-0398}  \and \\
Hyun-seok Min\inst{3}\orcidlink{0000-0002-7435-7884}  \and
Sungchul Choi\inst{1}\orcidlink{0000-0002-5836-3838} }

\authorrunning{Jung et al.}

\institute{Major in Industrial Data Science \& Engineering, \\ Department of Industrial and Data Engineering,  Pukyong National University \\ \email{\{kkyuhun94,sod7050,jsw6872\}@pukyong.ac.kr,sc82.choi@pknu.ac.kr} \and Teamreboott Inc. \\ \email{jaeyoungkim@reboott.ai} \and Tomocube Inc.\\ \email{hsmin@tomocube.com}
}

\maketitle

\begin{abstract}
In this paper, we present an effective data augmentation framework leveraging the Large Language Model (LLM) and Diffusion Model (DM) to tackle the challenges inherent in data-scarce scenarios. 
Recently, DMs have opened up the possibility of generating synthetic images to complement a few training images. However, increasing the diversity of synthetic images also raises the risk of generating samples outside the target distribution. Our approach addresses this issue by embedding novel semantic information into text prompts via LLM and utilizing real images as visual prompts, thus generating semantically rich images. To ensure that the generated images remain within the target distribution, we dynamically adjust the guidance weight based on each image's CLIPScore to control the diversity. Experimental results show that our method produces synthetic images with enhanced diversity while maintaining adherence to the target distribution. Consequently, our approach proves to be more efficient in the few-shot setting on several benchmarks. Our code is available at \href{https://github.com/kkyuhun94/dalda}{\color{magenta}https://github.com/kkyuhun94/dalda}.


\end{abstract}

\section{Introduction}
\label{sec:intro}
The emergence of Diffusion Models (DMs)~\cite{ddpm, ddim, imagen, stablediffusion, glide} has significantly increased research interest in potential applications of synthetic images, particularly in the field of data augmentation. Recent studies have shown that synthetic images generated by DMs can effectively complement real data~\cite{issynthetic, dafusion, azizi2023synthetic}. However, enhancing the diversity of synthetic images while remaining within the target distribution is still a challenge~\cite{scalinglaws, diversitydiffusion,synrobust}. For classification tasks, synthetic images produced by DMs are effective in extending semantic representation for in-distribution data, but they can also generate images that fall outside the target distribution~\cite{synrobust}. Particularly, as diversity increases, it can lead to the generation of samples that do not retain invariant characteristics for target classes~\cite{diversitydiffusion}. These synthetic images can be noise samples for training a classifier, leading to sub-par performance.

Real Guidance~\cite{issynthetic} addresses this issue by using real images to guide the data synthesis process, thereby reducing domain shift. Specifically, it introduces real images into the diffusion process to align synthetic data more closely with the target domain, effectively narrowing the gap between synthetic and real data distributions. This approach has demonstrated that guiding the synthesis process with real image information is beneficial for data augmentation. Furthermore, DA-Fusion~\cite{dafusion} proposes a robust data augmentation technique for few-shot classification using Textual Inversion~\cite{textualinversion} to fine-tune DM on training images for each class. By modifying the appearance of objects while preserving their semantic invariances, DA-Fusion enhances the diversity of synthetic images and adapts the DM to new domains, leading to superior performance in few-shot scenarios. 

While the previous methods help the generated data to stay within the target distribution, they may excessively limit diversity. Additionally, these methods may struggle to supplement new semantic information beyond the real image, especially when using CLIP~\cite{clip} templates like \texttt{``a photo of a \{class\}''}. This can be effective in maintaining the consistency of synthetic images, but it can be challenging to supplement various patterns that can be expressed with its class~(see~\cref{fig:qa}). 

Meanwhile, several studies have explored enhancing text prompts to increase the diversity of suitable synthetic images~\cite{issynthetic, scalinglaws, diversitydiffusion}. Fan \etal~\cite{scalinglaws} compares the diversity and recognizability of synthetic images generated using various prompt methods and demonstrates that improving class-specific prompts aids in generating synthetic images. However, simply expanding text prompts can still lead to DMs generating class-inconsistent synthetic images. 
Marwood \etal~\cite{diversitydiffusion} emphasizes that carefully controlling the influence of prompts is crucial for creating synthetic images that are beneficial for data augmentations.

In this paper, we introduce the \textit{Data Augmentation Leveraging Diffusion Model and LLM with Adaptive Guidance Scaling} (DALDA), a novel method leveraging synthetic images to enhance data augmentation. Our approach begins by utilizing the capabilities of the LLM (GPT-4) to generate text prompts that include various descriptions, such as actions, viewpoints, and environmental features. By integrating dataset and class-specific information into the LLM, we produce unique prompts that retain the intrinsic characteristics of each class. This methodology enriches the semantic content, particularly where each class has limited image samples.
To further refine this process, we present Adaptive Guidance Scaling (AGS), a mechanism designed to balance the influence of example images and text prompts during image generation. Prior to generating synthetic images, we employ the CLIP model to calculate the CLIPScore~\cite{clipscore}, which quantifies the alignment between an example image and its corresponding class name. Based on this score, we dynamically adjust the weights of prompt guidance within the Multi-Modal conditional Diffusion Model (MMDM).
This ensures that synthetic images do not deviate from the target distribution and maintains the consistency of each class while generating diverse synthetic images.

Our framework effectively balances the trade-off between increasing the diversity of synthetic images and ensuring they remain consistent within the target distribution. We conduct experimental analyses to evaluate the impact of each component on the diversity of synthetic images and the performance of downstream classification models. Our results demonstrate that our approach achieves higher diversity and improved performance of downstream classification models with fewer shots compared to existing methods.
The main contributions of our work can be summarized as:
\begin{itemize}
    \item We propose a data augmentation framework that leverages LLM-generated text prompts, enriched with novel semantic information, to increase the diversity. Additionally, we analyze the impact of LLM-generated prompts on synthetic images created by DM from a data augmentation perspective.
    \item We introduce AGS, which utilizes CLIPScore to adaptively adjust the balance between text and image conditions during image generation. This ensures that the synthetic images remain within the target distribution while reflecting the enriched text prompts.
    \item DALDA generates diverse synthetic images using pre-trained DM without additional fine-tuning. It shows better results than existing methods in both diversity and classification accuracy in few-shot scenarios.
\end{itemize}

\section{Related Work}
\label{sec:relatedwork}

\subsection{Data Augmentation with Generative Models}
The advancements in generative models have led to a surge of interest in leveraging these models for data augmentation. In particular, text-to-image~(T2I) generation models such as Stable Diffusion~(SD)~\cite{stablediffusion} have shown exceptional capability in producing highly diverse synthetic images, making them promising tools for augmenting training data.
For instance, GenImage~\cite{genimage} has demonstrated the potential of using T2I DMs to generate synthetic clones that could effectively replace real images in certain contexts. This study emphasized that synthetic images generated solely from text prompts could serve as viable substitutes for real data.

Various fine-tuning techniques such as ControlNet~\cite{controlnet}, DreamBooth~\cite{dreambooth}, GLIDE~\cite{glide} and Textual Inversion~\cite{textualinversion} facilitate more controlled image generation using multi-modal prompts (\eg, image, sketch and segmentation mask), expanding the versatility and applicability of DMs in various tasks. For example, ControlNet allows precise control over the output by conditioning the model on additional input data such as sketches or segmentation maps, thereby enhancing the utility of synthetic images in downstream tasks.
ScribbleGen~\cite{scribblegen} develops a data augmentation framework using ControlNet with a scribble prompt, demonstrating that this multi-modal prompt is helpful in data augmentation.
However, despite these advancements, previous data augmentation approaches often require fine-tuning DMs on sufficient training data to generate appropriate synthetic images.

Real Guidance~\cite{issynthetic} utilizes GLIDE~\cite{glide}, guided by a small number of real images, to generate synthetic images that mitigate domain variations. 
DA-Fusion~\cite{dafusion} introduces a framework for few-shot classification using Textual Inversion~\cite{textualinversion} that employs image-to-image transformations parameterized by pre-trained T2I DM. However, this method also necessitates fine-tuning DM. 
While these methods reduce domain shift and ensure class consistency, they generate synthetic images that are highly dependent on real images, which can be less effective in data-scarce scenarios.

Recent studies~\cite{unicontrolnet, t2iadapter, ipadapter} have explored fine-tuning pre-trained DMs through learnable adapters, enabling the creation of more controllable and diverse synthetic images. In our approach, we adopt~\cite{ipadapter}, which focuses on preserving instance characteristics while allowing for diverse variations.

\subsection{Automated Data Augmentation}

Traditional data augmentation methods increase the diversity of data by manipulating or transforming images (\eg, flip, crop, rotate, and blur). However, these methods require manually finding suitable techniques based on the distribution and patterns of the dataset. To overcome these limitations, automated data augmentation techniques such as AutoAugment~\cite{autoaugment} and RandAugment~\cite{randaugment} have been developed. AutoAugment uses reinforcement learning to find the optimal augmentation policies by constructing a search space of various augmentation methods, probabilities, and intensities. RandAugment reduces the search space by applying the same transformation intensity to all augmentation methods and optimizing only the number of augmentation techniques.

However, these approaches cannot add new semantic information through simple transformation. To extend automated data augmentation with DM-based techniques, we propose a strategy that extracts features from the original image and automatically determines the weights for the image prompts in MMDM.

\subsection{Class-Specific Prompts Augmentation}
\label{sec:promptsaugmentation}
To utilize DMs more effectively, recent studies have focused on the composition of text prompts.
He \etal~\cite{issynthetic} proposes Language Enhancement that extends class labels into sentence structures using T5~\cite{t5} for text prompt enhancement. It enhanced the diversity of images generated by DM and improved the effectiveness of synthetic data.
Additionally, Fan \etal~\cite{scalinglaws} shows the impact of synthetic images on classification tasks using various text prompts. 
This study observes the effectiveness of several types of text prompts, including simple class names like \texttt{``dog''}, class names with detailed descriptions from WordNet~\cite{wordnet}, and class names with hypernyms. It also examined CLIP templates, Word2Sentence using T5 to change the class names into a sentence~(\eg, \texttt{``a dog on the sofa''}), and image captioning (BLIP2~\cite{blip2}). Among these, Image captioning achieves the highest Top-1 accuracy on ImageNet~\cite{imagenet}. However, this method simply generates prompts by captioning the original images, making it difficult to add novel semantic information and resulting in lower diversity compared to CLIP templates and Word2Sentence.

Marwood \etal~\cite{diversitydiffusion} analyzes the potential and limitations of synthetic images in T2I systems, as well as the relationship between synthetic images and text prompts. This study demonstrates that synthetic images exhibit a trade-off between diversity and consistency depending on the influence of text prompts. This underscores the importance of carefully controlling the influence of prompts.
In this paper, we first provide the LLM with target data information and class details, instructing it to generate class-specific prompts enriched with additional semantic information. 

\begin{figure}[tb]
    \centering
    \includegraphics[width=10cm]{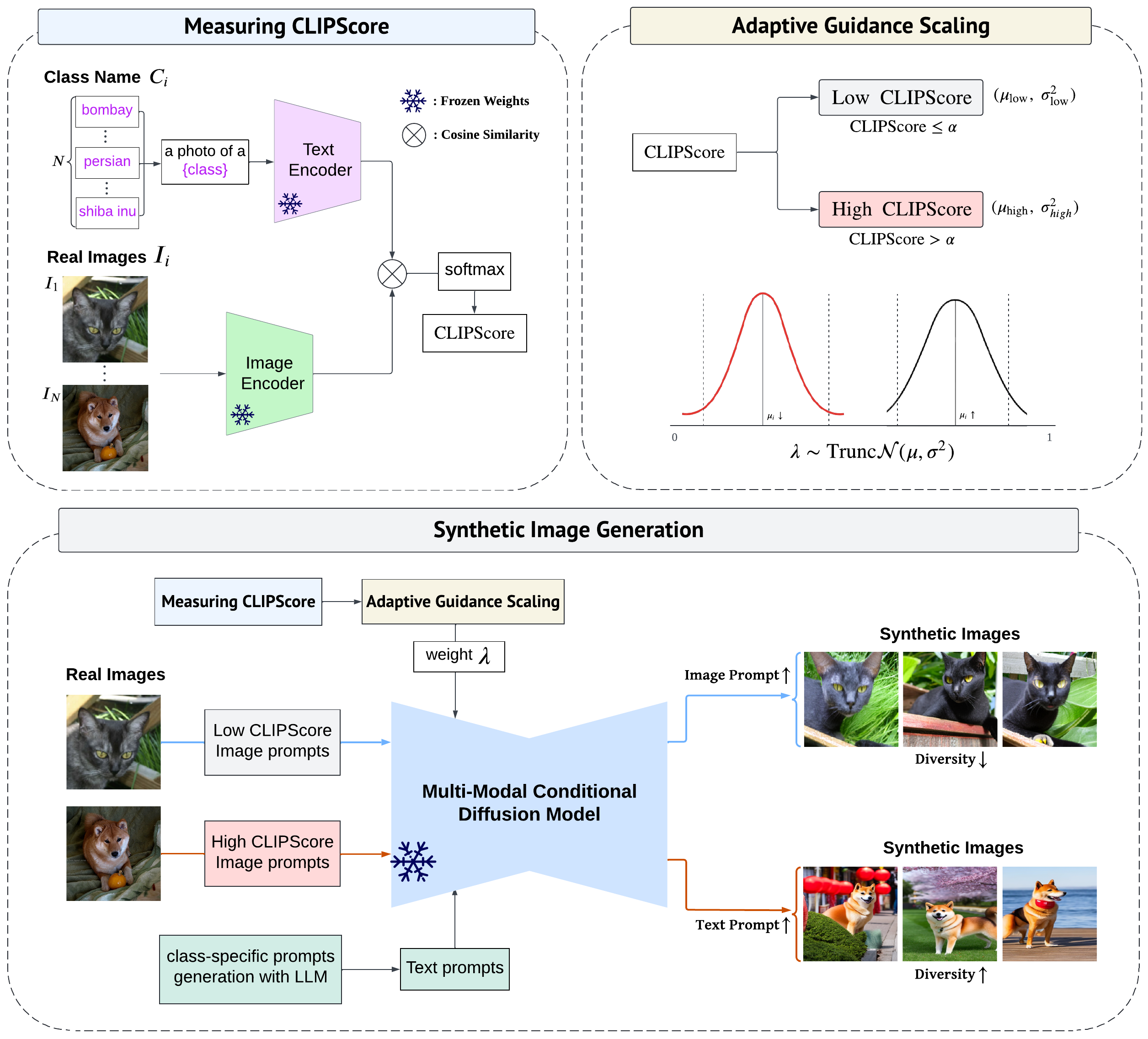}
    \caption{Overview of the proposed framework. We calculate the CLIPScore for each training image~(\cref{sec:imagescoring}). We then adaptively adjust the weight $\lambda$ of the prompt. $\lambda$ on samples with low CLIPScore serves to focus on the image guides and vice versa, weighting the text guides more heavily to obtain synthetic data with increased diversity~(\cref{subsec:ags}). Lastly, text prompts generated by LLM and image prompts are fed into MMDM to generate synthetic images~(\cref{sec:datagen}).} 
    \label{fig:framework}
\end{figure}

\section{Methods}
\label{sec:methods}

Our framework is illustrated in~\cref{fig:framework}. DALDA consists of multiple stages, including (1) measuring CLIPScore, (2) adaptive guidance scaling, and (3) synthetic data generation. Each of these stages will be explained in detail in the subsequent sections.

\subsection{Measuring CLIPScore}
\label{sec:imagescoring}
Before generating synthetic images with the MMDM, we calculate the CLIPScore~\cite{clipscore} which is the cosine similarity between the input image and its caption. In this study, we calculate the CLIPScore between the input image prompt and the text prompt \texttt{``a photo of a \{class\}''}. The CLIP model can be used to evaluate how well synthetic images reproduce the concept of a specific class~\cite{diversitydiffusion}.
In contrast, we measure the relationship between real images and class names and use this as a factor to adjust the influence of image prompts and text prompts.

For example, a low CLIPScore indicates that the real image and its class are not appropriately matched within the CLIP embedding. This suggests that, at the stage of generating synthetic images, even if the weight of the image prompt is decreased and the weight of the text prompt is increased, the MMDM is likely to fail in appropriately reflecting the text prompt~(see red box in~\cref{fig:qalowpets}). Conversely, a high CLIPScore indicates that the real image and class are well-matched, enabling the generation of synthetic images that accurately reflect the text prompt in the image prompt (see~\cref{fig:qahighpets}). 

We hypothesize that this phenomenon occurs because most open MMDMs, using SD as their backbone, utilize a CLIP encoder. When the multi-modal prompts do not align correctly between the image and text, it can lead to mismatches in the MMDMs' generation process, resulting in synthetic images that deviate from the target distribution.

\subsection{Adaptive Guidance Scaling}
\label{subsec:ags}

We adopt IP-Adapter~\cite{ipadapter} that specializes in generating synthetic images that maintain the characteristics of the image prompt while reflecting the text prompt. It employs a decoupled cross-attention mechanism with separated cross-attention layers for text features and image features. In this mechanism, the image features $c_i$ and text features $c_t$ from the CLIP encoders are passed through their respective cross-attention layers and then their outputs are added together.

The output of the decoupled cross-attention \( Z^{new} \) is defined as follows:

\begin{equation}
\begin{split}
Z^{new} = \text{Attention}(Q, K_t, V_t)+\lambda\cdot\text{Attention}(Q, K_i, V_i),
\end{split}
\label{eq:ipadapter}
\end{equation}
where \(\: Q=ZW_{q},\: K_t=c_tW_{kt},\: V_t=c_tW_{vt},\: K_i=c_iW_{ki},\: V_i=c_iW_{vi}\).
\begin{figure}[tb]
    \centering
    \includegraphics[width=12cm]{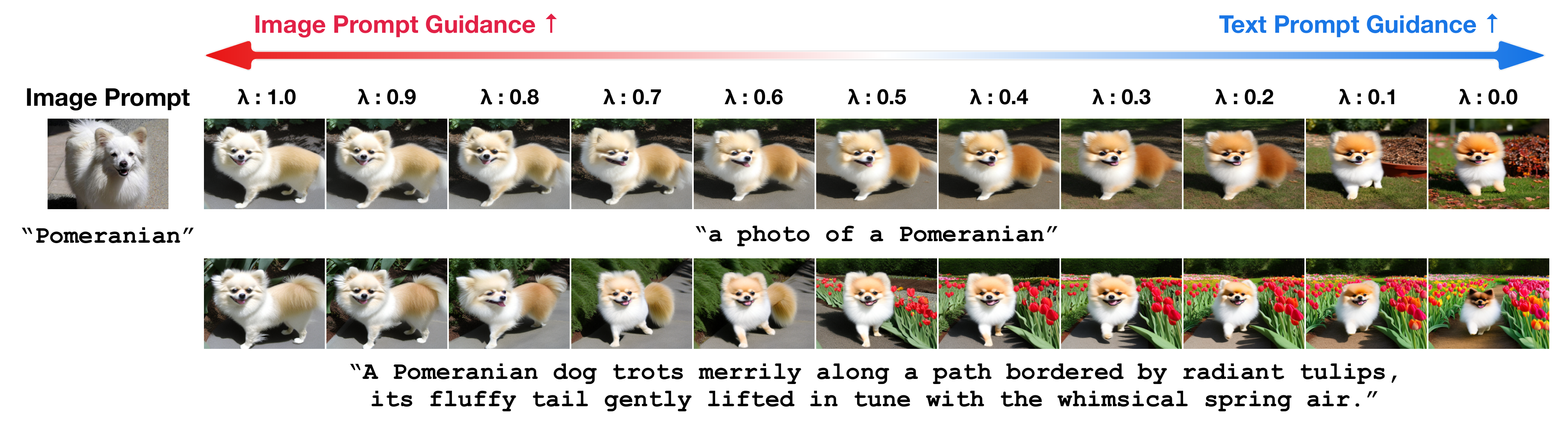}
    \caption{Example of synthetic images varying the $\lambda$. As the weight $\lambda$ gets closer to zero, it becomes more similar to T2I generation.}
    \label{fig:scaling}
\end{figure}
In~\cref{eq:ipadapter}, $Q$ is a matrix derived from query features $Z$ and weights $W_q$. $K_t$ and $V_t$ are matrices obtained from text features $c_t$ and their respective weights $W_{kt}$ and $W_{vt}$, while $K_i$ and $V_i$ are matrices derived from image features $c_i$ and their respective weights $W_{ki}$ and $W_{vi}$. The weight factor denoted by $\lambda$, assigns weights to image features. A higher value of $\lambda$ increases the weight of image guidance in the inference stage, while a lower value of $\lambda$ relatively increases the weight of text guidance~(\cref{fig:scaling}). 

Our framework dynamically adjusts the guidance weight $\lambda$ based on the CLIPScore of the example image during the synthetic image generation process. It can also be applied to other MMDMs by setting and using a parameter $\lambda$ to adjust the weights of image prompts or text prompts.

We measure the CLIPScore between each image prompt and class, then apply AGS to determine the direction of MMDM's image generation. AGS samples the $\lambda$ from a truncated normal distribution based on the CLIPScore for each real image. A truncated normal distribution is a variation of the normal distribution where values are bound within a specific range. This ensures that sampled values, in this case, the prompt guidance weight $\lambda$ do not exceed predefined limits. 

We categorize CLIPScore into two levels: High and Low, using a threshold $\alpha$ (default = 0.3). For image prompts with low CLIPScore~(Low), we restrict the prompt weight $\lambda$ to the range $[\text{min}_\text{Low}, \text{max}_\text{Low}]$ to reduce the influence of the text prompt and encourage variations closer to the image prompt. Conversely, for image prompts with high CLIPScore~(High), data augmentation can be more effective by increasing the influence of the text guidance. In this case, we restrict the prompt weight $\lambda$ to the range $[\text{min}_\text{High}, \text{max}_\text{High}]$.

In both scenarios, the sampling process is defined by a truncated normal distribution as follows:
\begin{equation}
\lambda \sim \text{Trunc}\mathcal{N}(\mu, \sigma^2, \text{min}, \text{max}),
\end{equation}
where the mean $\mu$ depends on the CLIPScore and is calculated as:
\begin{equation}
\mu = \text{min} + (\text{max} - \text{min}) \times (1 - \text{CLIPScore}).
\end{equation}
This way, higher CLIPScore results in lower $\lambda$ values, while lower CLIPScore leads to relatively higher $\lambda$ values being sampled. The reason for increasing the $\lambda$ for images with low CLIPScore is to mitigate the risk of MMDM generating images that deviate from the target distribution. While this adjustment may not yield significant performance improvements, it is deemed preferable to avoid generating incorrect samples that could adversely impact the results. Whereas, for images with high CLIPScore, the probability of deviating from the target distribution is low. Therefore, the weight of the text prompt is increased to enhance the diversity of the generated synthetic images.

As the number of examples per class increases, the effectiveness of synthetic images may decrease~\cite{issynthetic}. This is because the increased number of real images can compensate for a part of the required diversity. Additionally, as more real images are used, the risk of domain shift decreases, leading us to mitigate the limitations on diversity. To gradually relax the constraints on diversity as the number of real image examples increases, we set $\sigma = 0.05 \times  n$, where $n$ represents the number of examples per class. This strategy allows us to leverage the information of each image to maximize diversity while maintaining the semantic features of the class. 

\subsection{Synthetic Data Generation}
\label{sec:datagen}
\begin{figure}[tb]
    \centering
    \includegraphics[width=10cm]{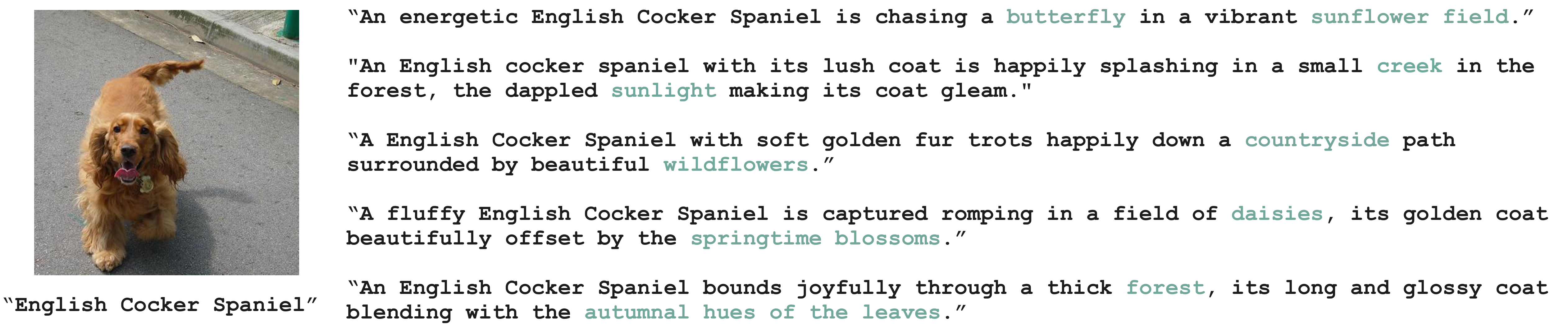}
    \caption{Example of text prompts generated by the LLM.}
    \label{fig:llm_prompt_examples}
\end{figure}

\noindent \textbf{Text Prompt Generation with LLM.}
To generate text prompts for the MMDM, we enhance the input by providing information about the target dataset and class details to the LLM. Additional instructions are added to increase the diversity of semantic content. For this purpose, we use the GPT-4 (gpt4-0613)~\cite{gpt4}.

In this stage, we generate multiple prompts for each sample, iterating $M$ times for $N$ training samples. Specifically, we incorporate descriptions of the downstream dataset and class information into the LLM. This aids the LLM in producing more natural and diverse prompts. Additionally, when generating prompts for a specific class, we supply the in-distribution classes to the LLM and include instructions to ensure the LLM avoided representing multiple in-distribution classes within a single prompt.

In~\cref{fig:llm_prompt_examples}, the generated text prompts show that the LLM can create class-centric sentences that are realistic, stay within the provided information, and maintain the semantic characteristics of each class while generating scenarios in various environments.
Details of the hyperparameters and actual prompts are reported in the supplementary material.

\noindent \textbf{Synthetic Image Generation.}
The resultant text and image prompts are randomly sampled and then fed into the MMDM. For each sample, we generate $M$ synthetic images using $M$ text prompts. In the 1-shot setting, the total number of synthetic data is $M\times N$. As the number of real examples per class increases, the total number of synthetic images becomes $M \times N$ multiplied by the number of real examples per class.
For downstream classification tasks, a classifier can be trained with both real and synthetic datasets.

\section{Experiments}
\label{sec:experiment}
\subsection{Setup}

\noindent \textbf{Datasets.}
\begin{figure}[tb]
     \centering
     \begin{subfigure}[b]{0.32\textwidth}
         \centering
         \includegraphics[width=\textwidth]{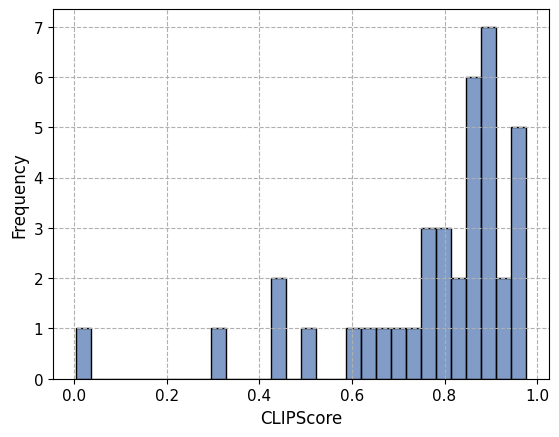}
         \caption{Oxford Pets~(HC)}
         \label{fig:pets}
     \end{subfigure}
     \begin{subfigure}[b]{0.32\textwidth}
         \centering
         \includegraphics[width=\textwidth]{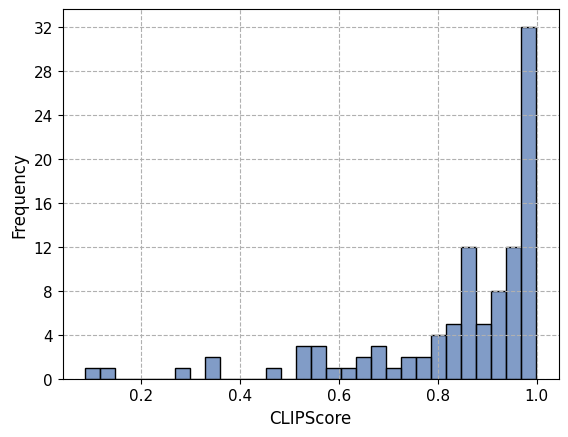}
         \caption{Caltech-101~(HC)}
         \label{fig:caltech}
     \end{subfigure}
     \begin{subfigure}[b]{0.32\textwidth}
         \centering
         \includegraphics[width=\textwidth]{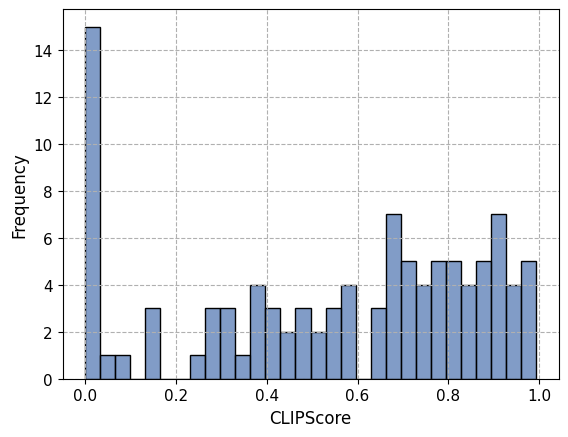}
         \caption{Flowers102~(LC)}
         \label{fig:flowers}
     \end{subfigure}
     \caption{CLIPScore distribution of datasets. Oxford Pets and Caltech-101 belong to the High CLIPScore (HC) group, with images from each class showing high CLIPScores. In contrast, Flowers102 has a higher proportion of classes with low CLIPScores, placing it in the Low CLIPScore (LC) group.}
     \label{fig:datasets}
\end{figure}
To demonstrate that our method can generate diverse synthetic images in data-scarce scenarios, we measure the accuracy in few-shot classification tasks.
We recognize the importance of accurate alignment between real images and class names in the generation of synthetic data. For a comprehensive and fair analysis, we select a subset of few-shot datasets from those used in~\cite{issynthetic}, based on their CLIPScores. These CLIPScores are calculated using the same methodology described in \cref{sec:imagescoring}.
Each dataset exhibits various distributions in~\cref{fig:datasets}; 
Caltech-101~\cite{caltech}, which belongs to the common dataset, recorded a high average CLIPScore of 0.8406, categorizing it as a High CLIPScore (HC) group. Oxford Pets~\cite{pets}, which belongs to the fine-grained dataset, also showed a relatively high CLIPScore of 0.7782, placing it in the HC group as well. In contrast, Flowers102~\cite{flowers}, another fine-grained dataset, has an average CLIPScore of 0.5548, placing it in the Low CLIPScore (LC) group.

\noindent \textbf{Implementation Details.}
For all experiments on classification tasks, we use \texttt{ResNet50}~\cite{resnet} pre-trained on ImageNet-1k and \texttt{CLIP-ViT-B/16}. For each training sample, the number of text prompts $M=10$. For training the classifier, we sample the real training images and the generated images with a uniform distribution.
More details on hyperparameters can be found~\cref{tab:hyperparameter} in Appendix~\ref{sec:appendixhyper}.

We compare our method with existing state-of-the-art methods, Real Guidance~\cite{issynthetic} and DA-Fusion~\cite{dafusion}. For these methods, we utilize the implementations found in~\cite{dafusion} to reproduce their methods. The hyperparameters are set according to the specifications provided in each respective paper. Base Prompt refers to using the CLIP template, \texttt{``a photo of a \{class\}''}, and LLM Prompt refers to using prompts generated by LLM. Random Scaling~(RS) refers to generating synthetic images by randomly scaling the $\lambda$ values in MMDM. For all quantitative experiments, we measure the average of three trials.

\subsection{Experimental Results}

\begin{table}[tb]
  \caption{Diversity analysis results on the 1-shot setting. DALDA achieves enhanced diversity compared to other methods. The best results are highlighted in bold.}
  \label{tab:diversity}
  \centering
  \resizebox{\textwidth}{!}{ 
  \begin{tabular}{@{}ccccccc@{}}
    \toprule
    & \multicolumn{2}{c}{Oxford Pets~(HC)} & \multicolumn{2}{c}{Caltech-101~(HC)} & \multicolumn{2}{c}{Flowers102~(LC)}\\
    Methods & CLIP-I(↓) & LPIPS(↑) & CLIP-I(↓) & LPIPS(↑) & CLIP-I(↓) & LPIPS(↑)\\
    \midrule
    Real Guidance & 0.9450 & 0.3886 & 0.9272 & 0.3426 & 0.9523 & 0.4194\\
    DA-Fusion & 0.9105 & 0.4978 & 0.8793 & 0.4633 & 0.9271 & 0.5284\\
    \midrule
    Base Prompt + RS  & 0.9196 & 0.6744 & 0.8850 & 0.6866 & 0.9168 & 0.7492\\
    Base Prompt + AGS & 0.9229 & 0.6967 & 0.8978 & 0.6945 & 0.9297 & {\bf 0.7505}\\
    LLM Prompt + RS & 0.8947 & 0.6912 & {\bf 0.8708} & 0.6951 & {\bf 0.9161} & 0.7474\\
    {\bf LLM Prompt + AGS (DALDA)} & {\bf 0.8816} & {\bf 0.7090} & 0.8794 & {\bf 0.7050} & 0.9208 & 0.7459\\
  \bottomrule
  \end{tabular}
}
\end{table}

\begin{table}[tb]
  \caption{1-shot classification accuracy of classifiers trained on synthetic images.}
  \label{tab:accuracy}
  \centering
  \resizebox{\textwidth}{!}{ 
  \begin{tabular}{@{}ccccccc@{}}
    \toprule
    & \multicolumn{2}{c}{Oxford Pets~(HC)} & \multicolumn{2}{c}{Caltech-101~(HC)} & \multicolumn{2}{c}{Flowers102~(LC)}\\
    Methods & Acc-CLIP & Acc-RN50 & Acc-CLIP & Acc-RN50 & Acc-CLIP & Acc-RN50\\
    \midrule
    Real Guidance & 0.8623 & 0.6654 & 0.9182 & 0.7010 & 0.8176 & 0.5038\\
    DA-Fusion & 0.8662 & 0.6324 & 0.9159 & 0.6851 & {\bf 0.8333} & 0.5259\\
    \midrule
    Base Prompt + RS & 0.8715 & 0.7773 & 0.9148 & 0.7829 & 0.8109 & 0.5280\\
    Base Prompt + AGS & 0.8728 & 0.7891 & 0.9176 & {\bf 0.7916} & 0.8140 & 0.5212\\
    LLM Prompt + RS & 0.8742 & 0.8017 & 0.9184 & 0.7808 & 0.8285 & {\bf 0.5348}\\
    {\bf LLM Prompt + AGS (DALDA)} & {\bf 0.8745} & {\bf 0.8069} & {\bf 0.9202} & 0.7879 & 0.8189 & 0.5208\\
  \bottomrule
  \end{tabular}
  }
\end{table}

In this section, we present a comprehensive analysis of the diversity and efficacy of synthetic images generated by DALDA, across different datasets. Our evaluation utilizes well-established metrics, CLIP-I~\cite{dreambooth} and LPIPS~\cite{lpips} to quantify diversity. For CLIP-I, a lower score indicates greater diversity, while for LPIPS, a higher score indicates greater diversity.

\cref{tab:diversity} shows that the diversity of synthetic images generated by our method is generally higher compared to Real Guidance and DA-Fusion. This indicates that our method can supplement diversity with only one image per class compared to existing methods.
In the HC dataset, DALDA shows outstanding performance, demonstrating its ability to generate synthetic images that are diverse and rich in information. This underscores its utility as an invaluable resource for training classifiers, especially when data availability is constrained. However, as detailed in \cref{subsec:ags}, when dealing with image prompts with low CLIPScore, the strategy shifts towards minimizing diversity to mitigate the risk of generating inaccurate samples. Consequently, in the LC dataset, DALDA utilizing LLM Prompt and AGS tends to limit diversity more than other methods applying RS. This cautious approach ensures that the generated sample closely fits the target distribution, aiming to maintain alignment with the target distribution, even if it does not always lead to higher accuracy.

Additionally, we measure the accuracy of classifiers trained on synthetic images, as shown in~\cref{tab:accuracy}.  While DALDA does not outperform in the LC dataset, it consistently shows robust performance and achieves remarkable results in the HC dataset without additional fine-tuning. Additional statistical tests are provided in \cref{tab:acc} of \cref{sec:appendixstatic}.

\begin{figure}[tb]
     \centering
      \begin{subfigure}[b]{0.45\textwidth}
         \centering
         \includegraphics[width=\textwidth]{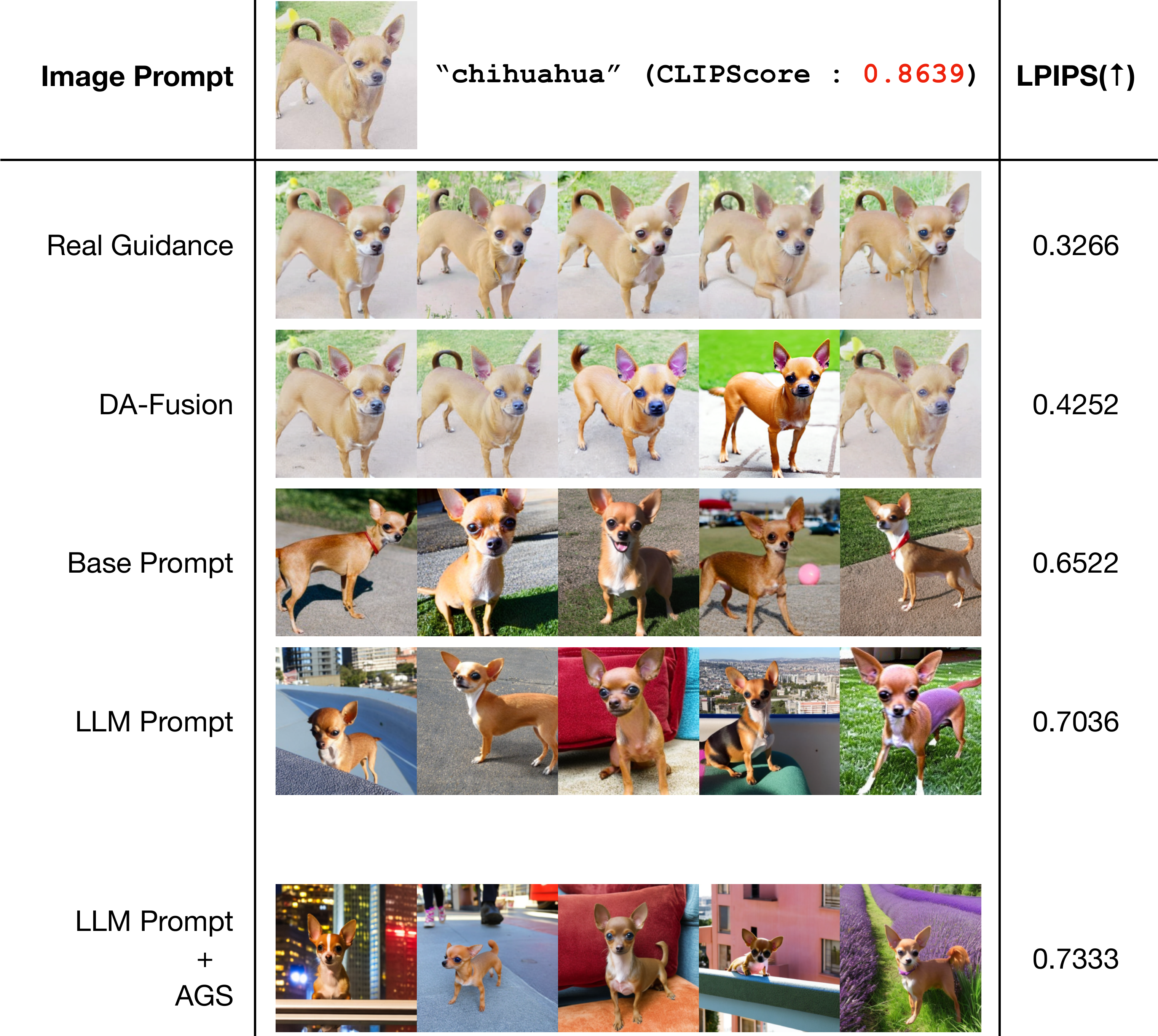}
         \caption{High CLIPScore Examples}
         \label{fig:qahighpets}
      \end{subfigure}
      \hfill
     \begin{subfigure}[b]{0.45\textwidth}
         \centering
         \includegraphics[width=\textwidth]{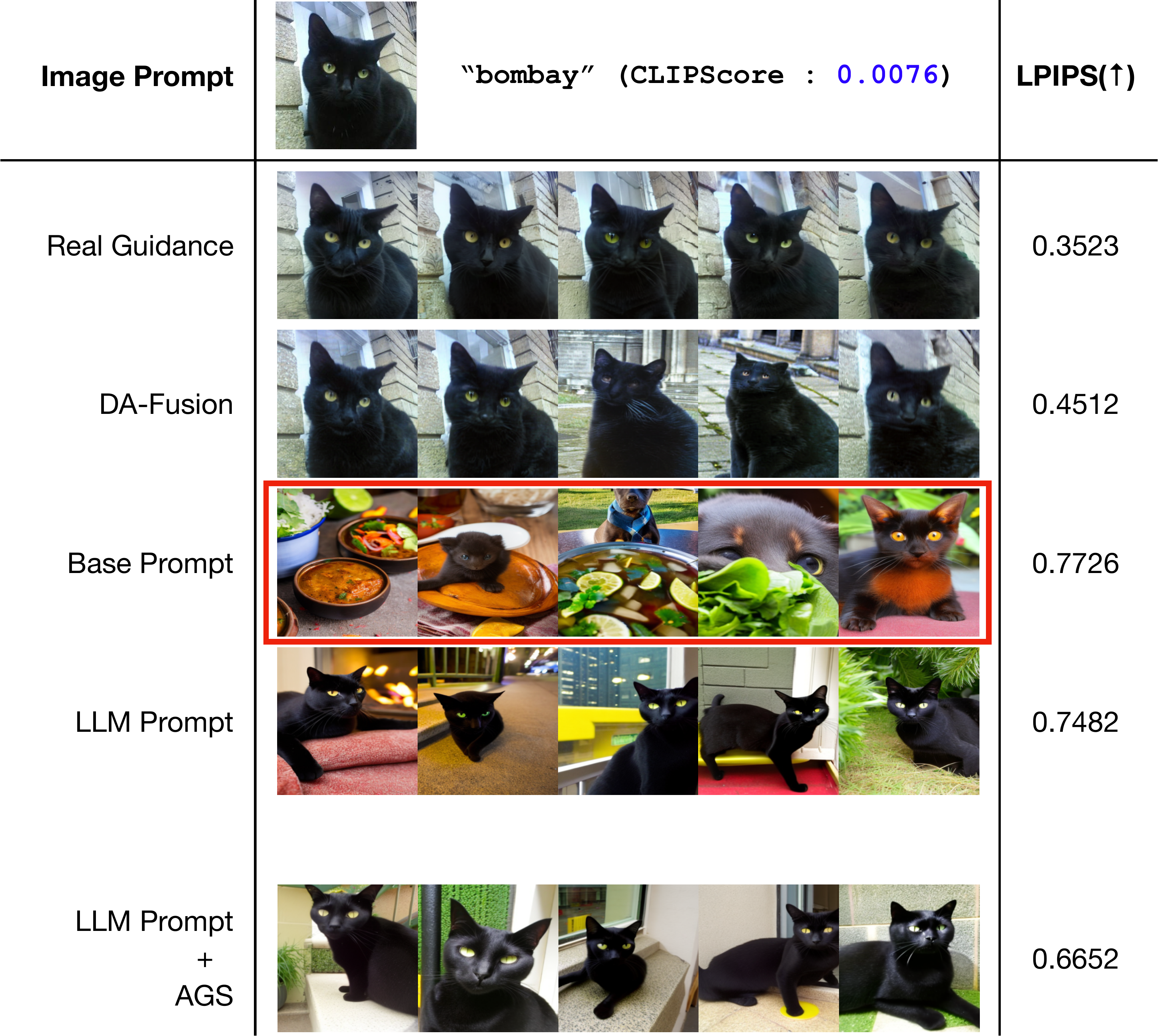}
         \caption{Low CLIPScore Examples}
         \label{fig:qalowpets}
     \end{subfigure}
     \caption{Synthetic image comparisons. Base Prompt refers to using the CLIP template, \texttt{``a photo of a \{class\}''}, and LLM Prompt refers to using prompts generated by the LLM. In the case of a low CLIPScore example, a Base Prompt might fail to maintain class consistency~(red box), resulting in the generation of synthetic images that do not accurately represent the intended class.}
     \label{fig:qa}
\end{figure}

In~\cref{fig:qa}, we conduct a qualitative comparison of our method, including the baselines. The visualization results analyze how synthetic images are generated for example images with low and high CLIPScore, demonstrating the effectiveness of our method in each case. Both Real Guidance and DA-Fusion fall short of the other methods in terms of diversity across the examples. Our final method, LLM Prompt + AGS, exhibits the highest diversity in high CLIPScore examples compared to the existing methods, while it shows the lowest diversity in low CLIPScore examples, excluding the two prior methods. This visually demonstrates that our intended AGS is designed to produce images with greater diversity in high CLIPScore examples while reducing diversity in low CLIPScore examples. When using high CLIPScore examples as image prompts~(\cref{fig:qahighpets}), AGS increases the diversity of synthetic images. Conversely, when using low CLIPScore examples as image prompts~(\cref{fig:qalowpets}), AGS decreases diversity to keep the synthetic images within the target distribution.

\begin{figure}[tb]
     \centering
     \begin{subfigure}[b]{0.35\textwidth}
         \centering
         \includegraphics[width=\textwidth]{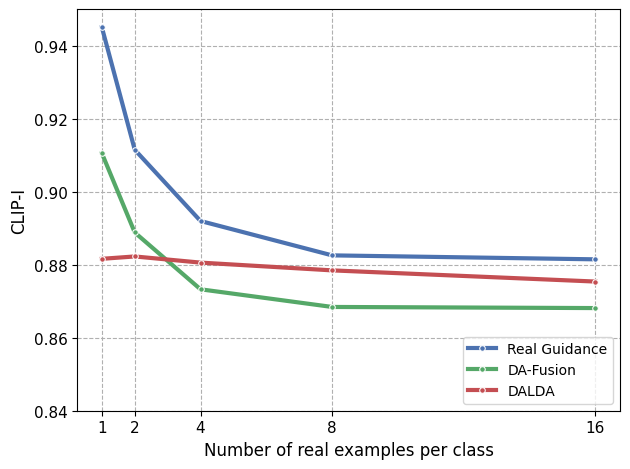}
         \caption{CLIP-I~($\downarrow$)}
         \label{fig:clipi}
     \end{subfigure}
     \begin{subfigure}[b]{0.35\textwidth}
         \centering
         \includegraphics[width=\textwidth]{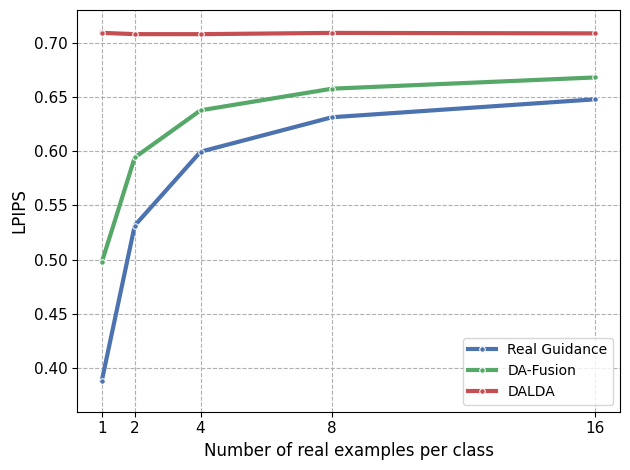}
         \caption{LPIPS~($\uparrow$)}
         \label{fig:lpis}
     \end{subfigure}
     \caption{Diversity comparison on Oxford Pets. Our method maintains a similar level of diversity even as the number of examples increases, compared to other methods. This demonstrates that our approach quickly converges to the target diversity even when there is only one example per class.}
     \label{fig:diversity16shot}
\end{figure}

\begin{figure}[tb]
     \centering
     \begin{subfigure}[b]{0.35\textwidth}
         \centering
         \includegraphics[width=\textwidth]{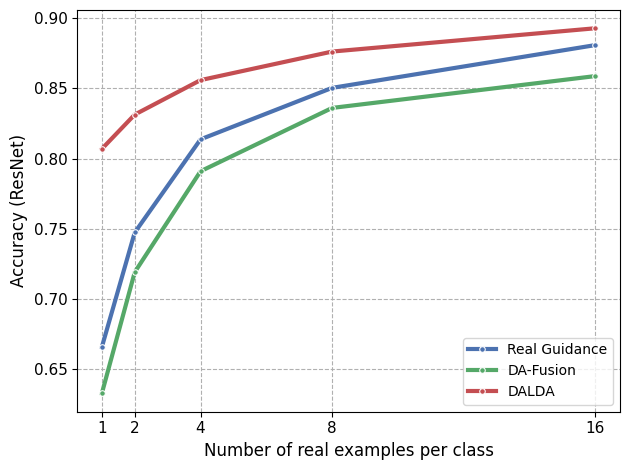}
         \caption{ResNet}
         \label{fig:petsrn50}
     \end{subfigure}
     \begin{subfigure}[b]{0.35\textwidth}
         \centering
         \includegraphics[width=\textwidth]{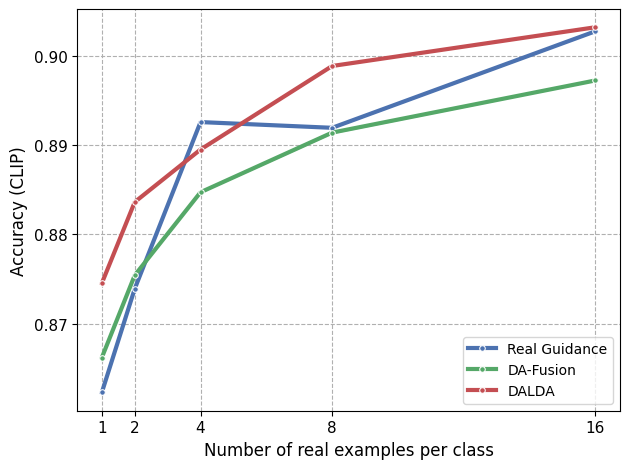}
         \caption{CLIP}
         \label{fig:petsclip}
     \end{subfigure}
     \caption{Comparison results with state-of-the-art approaches on the Oxford Pets shows that the proposed method achieves the highest N-shot accuracies in most settings. Our 1-shot accuracy with ResNet is comparable to the 4-shot accuracies of other methods.}
     \label{fig:fewshotpets}
\end{figure}

\begin{figure}[tb]
     \centering
     \begin{subfigure}[b]{0.32\textwidth}
         \centering
         \includegraphics[width=\textwidth]{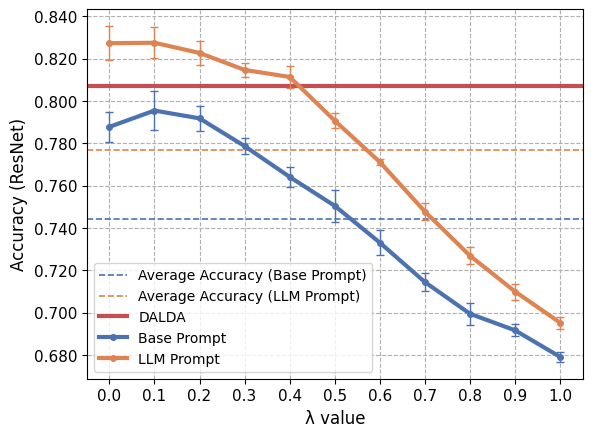}
         \caption{Oxford Pets~(ResNet)}
         \label{fig:promptrnpets}
     \end{subfigure}
     \begin{subfigure}[b]{0.32\textwidth}
         \centering
         \includegraphics[width=\textwidth]{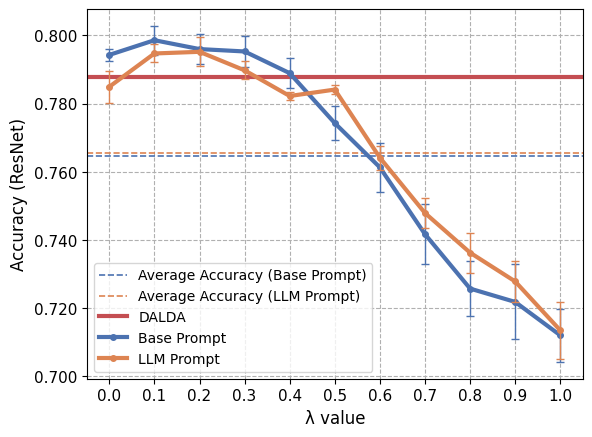}
         \caption{Caltech-101~(ResNet)}
         \label{fig:promptrncaltech}
     \end{subfigure}
          \begin{subfigure}[b]{0.32\textwidth}
         \centering
         \includegraphics[width=\textwidth]{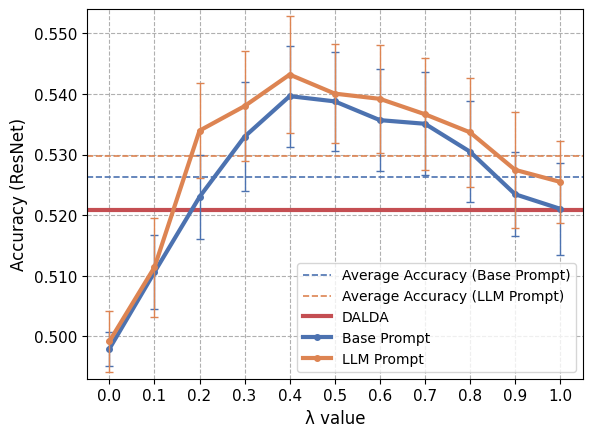}
         \caption{Flowers102~(ResNet)}
         \label{fig:promptrnflowers}
     \end{subfigure}
     \begin{subfigure}[b]{0.32\textwidth}
         \centering
         \includegraphics[width=\textwidth]{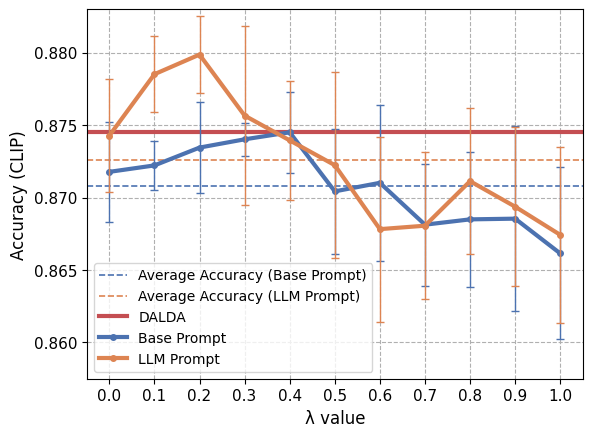}
         \caption{Oxford Pets~(CLIP)}
         \label{fig:promptCLIPpets}
     \end{subfigure}
     \begin{subfigure}[b]{0.32\textwidth}
         \centering
         \includegraphics[width=\textwidth]{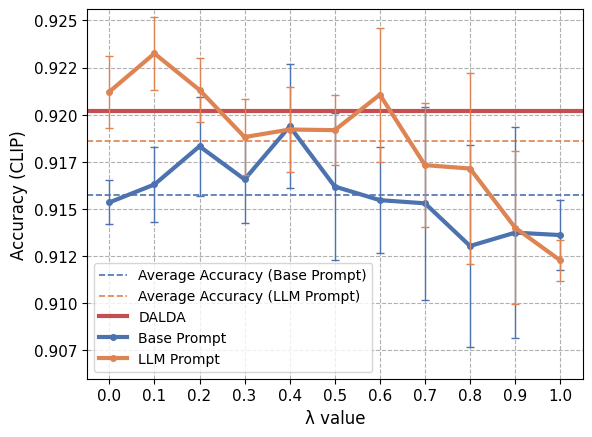}
         \caption{Caltech-101~(CLIP)}
         \label{fig:promptCLIPcaltech}
     \end{subfigure}
          \begin{subfigure}[b]{0.32\textwidth}
         \centering
         \includegraphics[width=\textwidth]{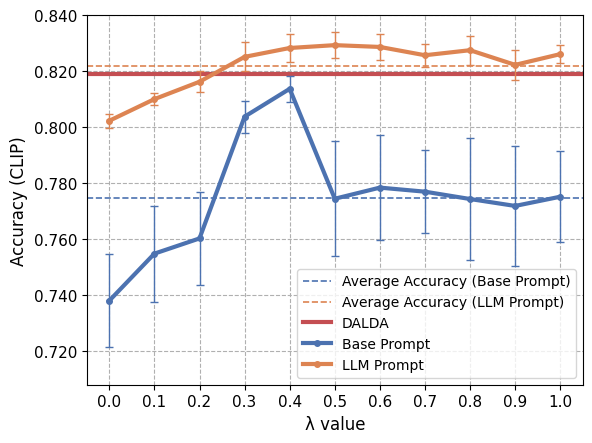}
         \caption{Flowers102~(CLIP)}
         \label{fig:promptCLIPflowers}
     \end{subfigure}
     \caption{LLM Prompt vs. Base Prompt~(1-shot accuracy). At a fixed prompt weight $\lambda$, LLM Prompt is generally more effective for model training compared to Base Prompt. Each horizontal line represents the average 1-shot accuracy for both the LLM and Base Prompt methods, as well as the accuracy achieved by DALDA.}
     \label{fig:prompteffect}
\end{figure}

\section{Analysis}

\noindent \textbf{Effectiveness of DALDA.}
We conducted comparative experiments~(\cref{fig:diversity16shot,fig:fewshotpets}) on Oxford Pets to evaluate the diversity and downstream models' accuracy when increasing the number of examples per class. Across all methods, the increase in the number of examples generally leads to a reduction in changes to diversity. In contrast, our method maintains a consistent level of diversity regardless of the number of examples. This indicates that our approach can quickly achieve the desired diversity with a small number of examples. For the CLIP-I score, our value falls between Real Guidance and DA-Fusion in~\cref{fig:clipi}. However, in terms of accuracy, our method surpasses both. This indicates that our approach not only focuses on enhancing diversity but also on limiting excessive diversity that could negatively affect the outcomes. Our method outperforms existing methods in terms of accuracy in~\cref{fig:fewshotpets}. In particular, with ResNet, our approach achieves similar or better results with just one example image compared to using four example images with other methods in~\cref{fig:petsrn50}.

\begin{figure}[tb]
     \centering
     \begin{subfigure}[b]{0.24\textwidth}
         \centering
         \includegraphics[width=\textwidth]{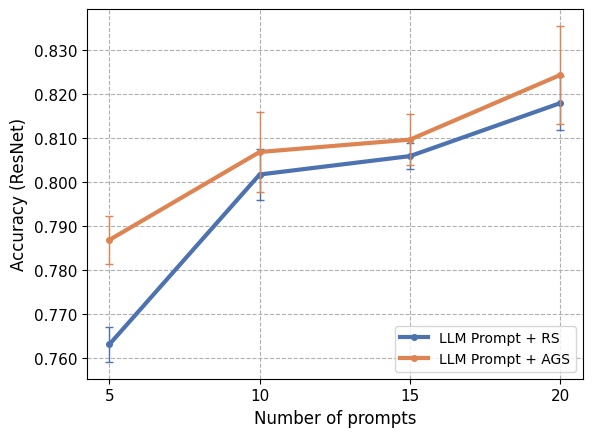}
         \caption{ResNet Accuracy}
         \label{fig:promptaccrn}
     \end{subfigure}
     \begin{subfigure}[b]{0.24\textwidth}
         \centering
         \includegraphics[width=\textwidth]{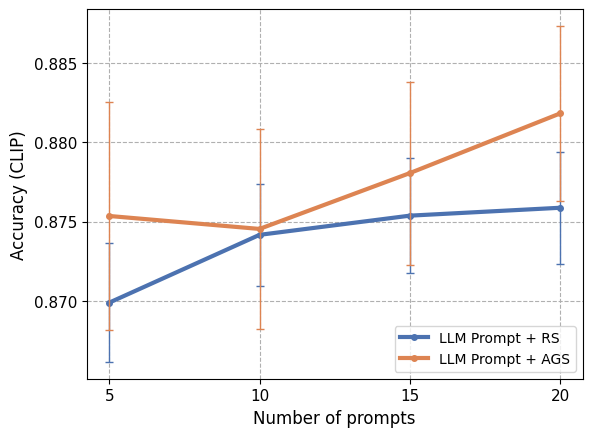}
         \caption{CLIP Accuracy}
         \label{fig:pormptaccclip}
     \end{subfigure}
     \begin{subfigure}[b]{0.24\textwidth}
         \centering
         \includegraphics[width=\textwidth]{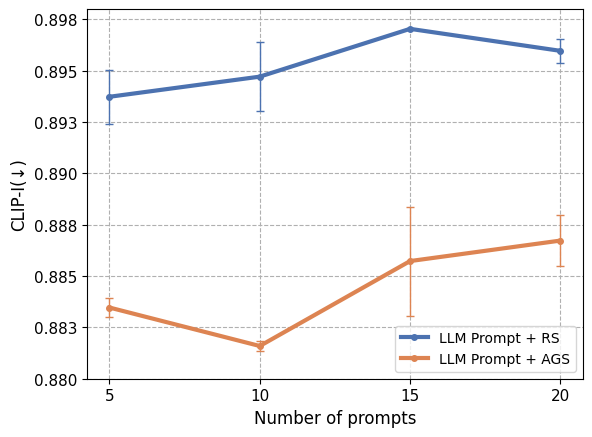}
         \caption{CLIP-I~($\downarrow$)}
         \label{fig:promptclipi}
     \end{subfigure}
     \begin{subfigure}[b]{0.24\textwidth}
         \centering
         \includegraphics[width=\textwidth]{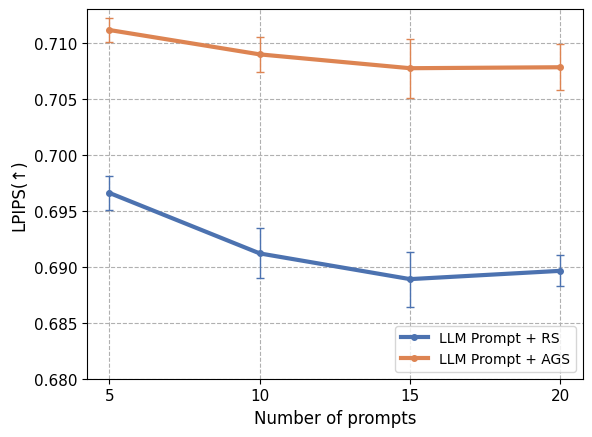}
         \caption{LPIPS~($\uparrow$)}
         \label{fig:promptlpips}
     \end{subfigure}
     \caption{Comparison experiments with scaling methods on Oxford Pets~(1-shot accuracy). Regardless of the number of prompts, in all cases where the LLM is used, AGS generates more diverse and effective synthetic images than RS.}
     \label{fig:numofprompt}
\end{figure}

\noindent \textbf{LLM Can Provide Beneficial Semantic Information.}
We conducted a comparative analysis of models trained on synthetic images generated by fixing $\lambda$ at increments of 0.1 from 0 to 1 according to two different text prompts formats~(\cref{fig:prompteffect}). Across all datasets, we found that using LLM Prompt generally yielded higher performance than using the Base Prompt. This indicates that leveraging the expanded semantic information from the LLM is more effective for data augmentation than generating synthetic images with a Base Prompt using MMDM. However, even with the LLM, the performance can be lower than the Base Prompt depending on how the prompt weight is utilized. This underscores the necessity of controlling for diversity to ensure it aligns with the purpose of data augmentation. In~\cref{fig:numofprompt}, we analyzed the impact of the number of prompts on Oxford Pets. Even as the number of LLM prompts increases, diversity does not change significantly. However, the accuracy of downstream models increases linearly with the number of prompts. This suggests that semantic diversity, in addition to visual diversity, influences model performance. It also demonstrates that AGS can produce more effective synergies when using LLM prompts.

\noindent \textbf{Fixing Guidance weight is Vulnerable to Distribution Changes.}
In~\cref{fig:prompteffect}, we found that fixing the $\lambda$ value according to the dataset can significantly impact the performance of synthetic images. Particularly, fixing the $\lambda$ value for all classes can make it dependent on the overall CLIPScore distribution of the dataset. For example, in the cases of Oxford Pets and Caltech-101 (HC), the best performance is achieved at lower $\lambda$ values, which reflect more on the text prompts. On the other hand, for Flowers102 (LC), it is difficult to find a linear relationship with changes in the $\lambda$, and even at lower $\lambda$ values, the accuracy for both Base and LLM Prompts is significantly lower than the average values. Therefore, it is important to recognize that fixing the $\lambda$ value or guidance weight according to the dataset can significantly degrade performance under certain conditions, emphasizing the need to adjust these parameters to suit the characteristics of the data. This is demonstrated by the values represented by each horizontal line in \cref{fig:prompteffect}. In the LC dataset, the accuracy of DALDA may not be significantly higher than the average accuracy of the Base Prompt and LLM Prompt. However, in the HC dataset, DALDA achieves higher accuracy compared to the average accuracy of the Base Prompt and LLM Prompt.

\section{Limitations and Future Works}
Our approach starts by measuring the CLIPScore of example images. This implies that our method, which adjusts the generation strategy based on CLIPScore, may inherently reflect the limitations of CLIP itself. This is especially true for LC datasets like Flowers102, where effectiveness is limited. Recent studies have investigated metrics that leverage feature values extracted from the embeddings of models~\cite{dinov2} pre-trained using self-supervised methods. Additionally, Hu \etal~\cite{tifa} proposed to evaluate how faithfully generated images adhere to text inputs using visual question answering. Exploring better metrics beyond CLIPScore could be a valuable research direction.

We evaluate synthetic images using image diversity metrics and the few-shot accuracy. While our method shows more pronounced performance improvements with ResNet, the performance improvements are relatively smaller for the CLIP classifier. This is likely because CLIP already includes a certain level of semantic information. This implies that the distribution of effective synthetic images may vary depending on the downstream model. Further exploring synthetic image generation strategies with downstream model knowledge could be beneficial.

\section{Conclusions}
We propose a framework, DALDA, that effectively utilizes LLM and DM for data augmentation in data-scarce scenarios. Our method aims to generate synthetic images that maintain class consistency while enriching diversity, even with just one example per class. By leveraging the knowledge of LLM, we create class-specific text prompts that introduce novel semantic information. We then apply AGS, dynamically adjusting the image generation strategy based on each image's CLIPScore. 
Our method enhances the diversity of synthetic images and improves downstream model performance in experiments. In 1-shot classification, DALDA achieved a mean accuracy improvement of 8.18\% over RG and 9.08\% over DA-Fusion. Through ablation studies, we provide insights by analyzing the components of our framework from a data augmentation perspective. This demonstrates that our approach is a powerful data augmentation method capable of achieving appropriate diversity in data-scarce situations.

\section*{Acknowledgement}
This work was supported by the IITP (Institute of Information \& Communications Technology Planning \& Evaluation)-ICAN (ICT Challenge and Advanced Network of HRD) (IITP-2024-RS-2023-00259806, 20\%) grant funded by the Korea government (Ministry of Science and ICT) and the National Research Foundation of Korea (NRF) grant funded by the Korea government (MSIT) (No. RS-2024-00354675, 50\% and No. RS-2024-00352184, 30\%).

\bibliographystyle{splncs04}
\bibliography{main}

\begin{thebibliography}{10}
\providecommand{\url}[1]{\texttt{#1}}
\providecommand{\urlprefix}{URL }
\providecommand{\doi}[1]{https://doi.org/#1}

\bibitem{gpt4}
Achiam, J., Adler, S., Agarwal, S., Ahmad, L., Akkaya, I., Aleman, F.L., Almeida, D., Altenschmidt, J., Altman, S., Anadkat, S., et~al.: Gpt-4 technical report. arXiv preprint arXiv:2303.08774  (2023)

\bibitem{azizi2023synthetic}
Azizi, S., Kornblith, S., Saharia, C., Norouzi, M., Fleet, D.J.: Synthetic data from diffusion models improves imagenet classification. Transactions on Machine Learning Research  (2023)

\bibitem{autoaugment}
Cubuk, E.D., Zoph, B., Mane, D., Vasudevan, V., Le, Q.V.: Autoaugment: Learning augmentation strategies from data. In: Proceedings of the IEEE/CVF conference on Computer Vision and Pattern Recognition. pp. 113--123 (2019)

\bibitem{randaugment}
Cubuk, E.D., Zoph, B., Shlens, J., Le, Q.V.: Randaugment: Practical automated data augmentation with a reduced search space. In: Advances in Neural Information Processing Systems. vol.~33, pp. 18613--18624 (2020)

\bibitem{imagenet}
Deng, J., Dong, W., Socher, R., Li, L.J., Li, K., Fei-Fei, L.: Imagenet: A large-scale hierarchical image database. In: 2009 IEEE Conference on Computer Vision and Pattern Recognition. pp. 248--255. IEEE (2009)

\bibitem{scalinglaws}
Fan, L., Chen, K., Krishnan, D., Katabi, D., Isola, P., Tian, Y.: Scaling laws of synthetic images for model training... for now. In: Proceedings of the IEEE/CVF Conference on Computer Vision and Pattern Recognition. pp. 7382--7392 (2024)

\bibitem{caltech}
Fei-Fei, L., Fergus, R., Perona, P.: One-shot learning of object categories. IEEE transactions on pattern analysis and machine intelligence  \textbf{28}(4),  594--611 (2006)

\bibitem{textualinversion}
Gal, R., Alaluf, Y., Atzmon, Y., Patashnik, O., Bermano, A.H., Chechik, G., Cohen-Or, D.: An image is worth one word: Personalizing text-to-image generation using textual inversion. In: The Eleventh International Conference on Learning Representations (2023)

\bibitem{resnet}
He, K., Zhang, X., Ren, S., Sun, J.: Deep residual learning for image recognition. In: Proceedings of the IEEE Conference on Computer Vision and Pattern Recognition. pp. 770--778 (2016)

\bibitem{issynthetic}
He, R., Sun, S., Yu, X., Xue, C., Zhang, W., Torr, P., Bai, S., QI, X.: Is synthetic data from generative models ready for image recognition? In: The Eleventh International Conference on Learning Representations (2023)

\bibitem{clipscore}
Hessel, J., Holtzman, A., Forbes, M., Le~Bras, R., Choi, Y.: Clipscore: A reference-free evaluation metric for image captioning. In: Proceedings of the 2021 Conference on Empirical Methods in Natural Language Processing. pp. 7514--7528 (2021)

\bibitem{ddpm}
Ho, J., Jain, A., Abbeel, P.: Denoising diffusion probabilistic models. Advances in Neural Information Processing Systems  \textbf{33},  6840--6851 (2020)

\bibitem{tifa}
Hu, Y., Liu, B., Kasai, J., Wang, Y., Ostendorf, M., Krishna, R., Smith, N.A.: Tifa: Accurate and interpretable text-to-image faithfulness evaluation with question answering. In: Proceedings of the IEEE/CVF International Conference on Computer Vision. pp. 20406--20417 (2023)

\bibitem{blip2}
Li, J., Li, D., Savarese, S., Hoi, S.: Blip-2: Bootstrapping language-image pre-training with frozen image encoders and large language models. In: International conference on machine learning. pp. 19730--19742. PMLR (2023)

\bibitem{diversitydiffusion}
Marwood, D., Baluja, S., Alon, Y.: Diversity and diffusion: Observations on synthetic image distributions with stable diffusion. arXiv preprint arXiv:2311.00056  (2024)

\bibitem{wordnet}
Miller, G.A.: Wordnet: a lexical database for english. Communications of the ACM  \textbf{38}(11),  39--41 (1995)

\bibitem{t2iadapter}
Mou, C., Wang, X., Xie, L., Wu, Y., Zhang, J., Qi, Z., Shan, Y.: T2i-adapter: Learning adapters to dig out more controllable ability for text-to-image diffusion models. In: Proceedings of the AAAI Conference on Artificial Intelligence. pp. 4296--4304 (2024)

\bibitem{glide}
Nichol, A.Q., Dhariwal, P., Ramesh, A., Shyam, P., Mishkin, P., Mcgrew, B., Sutskever, I., Chen, M.: Glide: Towards photorealistic image generation and editing with text-guided diffusion models. In: International Conference on Machine Learning. pp. 16784--16804. PMLR (2022)

\bibitem{flowers}
Nilsback, M.E., Zisserman, A.: Automated flower classification over a large number of classes. In: 2008 Sixth Indian conference on computer vision, graphics \& image processing. pp. 722--729. IEEE (2008)

\bibitem{dinov2}
Oquab, M., Darcet, T., Moutakanni, T., Vo, H.V., Szafraniec, M., Khalidov, V., Fernandez, P., HAZIZA, D., Massa, F., El-Nouby, A., Assran, M., Ballas, N., Galuba, W., Howes, R., Huang, P.Y., Li, S.W., Misra, I., Rabbat, M., Sharma, V., Synnaeve, G., Xu, H., Jegou, H., Mairal, J., Labatut, P., Joulin, A., Bojanowski, P.: {DINO}v2: Learning robust visual features without supervision. Transactions on Machine Learning Research  (2024)

\bibitem{pets}
Parkhi, O.M., Vedaldi, A., Zisserman, A., Jawahar, C.: Cats and dogs. In: 2012 IEEE conference on Computer Vision and Pattern Recognition. pp. 3498--3505. IEEE (2012)

\bibitem{clip}
Radford, A., Kim, J.W., Hallacy, C., Ramesh, A., Goh, G., Agarwal, S., Sastry, G., Askell, A., Mishkin, P., Clark, J., et~al.: Learning transferable visual models from natural language supervision. In: International Conference on Machine Learning. pp. 8748--8763. PMLR (2021)

\bibitem{t5}
Raffel, C., Shazeer, N., Roberts, A., Lee, K., Narang, S., Matena, M., Zhou, Y., Li, W., Liu, P.J.: Exploring the limits of transfer learning with a unified text-to-text transformer. Journal of Machine Learning Research  \textbf{21}(140),  1--67 (2020)

\bibitem{stablediffusion}
Rombach, R., Blattmann, A., Lorenz, D., Esser, P., Ommer, B.: High-resolution image synthesis with latent diffusion models. In: Proceedings of the IEEE/CVF conference on Computer Vision and Pattern Recognition. pp. 10684--10695 (2022)

\bibitem{dreambooth}
Ruiz, N., Li, Y., Jampani, V., Pritch, Y., Rubinstein, M., Aberman, K.: Dreambooth: Fine tuning text-to-image diffusion models for subject-driven generation. In: Proceedings of the IEEE/CVF Conference on Computer Vision and Pattern Recognition. pp. 22500--22510 (2023)

\bibitem{imagen}
Saharia, C., Chan, W., Saxena, S., Li, L., Whang, J., Denton, E.L., Ghasemipour, K., Gontijo~Lopes, R., Karagol~Ayan, B., Salimans, T., et~al.: Photorealistic text-to-image diffusion models with deep language understanding. Advances in Neural Information Processing Systems  \textbf{35},  36479--36494 (2022)

\bibitem{scribblegen}
Schnell, J., Wang, J., Qi, L., Hu, V.T., Tang, M.: Generative data augmentation improves scribble-supervised semantic segmentation. In: CVPR 2024 Workshop SyntaGen: Harnessing Generative Models for Synthetic Visual Datasets (2024)

\bibitem{synrobust}
Singh, K., Navaratnam, T., Holmer, J., Schaub-Meyer, S., Roth, S.: Is synthetic data all we need? benchmarking the robustness of models trained with synthetic images. In: CVPR 2024 Workshop SyntaGen: Harnessing Generative Models for Synthetic Visual Datasets (2024)

\bibitem{ddim}
Song, J., Meng, C., Ermon, S.: Denoising diffusion implicit models. In: International Conference on Learning Representations (2021)

\bibitem{dafusion}
Trabucco, B., Doherty, K., Gurinas, M., Salakhutdinov, R.: Effective data augmentation with diffusion models. In: The Twelfth International Conference on Learning Representations (2024)

\bibitem{ipadapter}
Ye, H., Zhang, J., Liu, S., Han, X., Yang, W.: Ip-adapter: Text compatible image prompt adapter for text-to-image diffusion models. arXiv preprint arXiv:2308.06721  (2023)

\bibitem{controlnet}
Zhang, L., Rao, A., Agrawala, M.: Adding conditional control to text-to-image diffusion models. In: Proceedings of the IEEE/CVF International Conference on Computer Vision. pp. 3836--3847 (2023)

\bibitem{lpips}
Zhang, R., Isola, P., Efros, A.A., Shechtman, E., Wang, O.: The unreasonable effectiveness of deep features as a perceptual metric. In: Proceedings of the IEEE conference on Computer Vision and Pattern Recognition. pp. 586--595 (2018)

\bibitem{unicontrolnet}
Zhao, S., Chen, D., Chen, Y.C., Bao, J., Hao, S., Yuan, L., Wong, K.Y.K.: Uni-controlnet: All-in-one control to text-to-image diffusion models. Advances in Neural Information Processing Systems  \textbf{36} (2024)

\bibitem{genimage}
Zhu, M., Chen, H., Yan, Q., Huang, X., Lin, G., Li, W., Tu, Z., Hu, H., Hu, J., Wang, Y.: Genimage: A million-scale benchmark for detecting ai-generated image. Advances in Neural Information Processing Systems  \textbf{36} (2024)

\end{thebibliography}

\clearpage

\appendix
\section*{Appendix}
\section{Statistical Test}
\label{sec:appendixstatic}
To provide a clearer analysis of the main results, we conducted additional statistical tests comparing our method with the baselines in \cref{tab:acc}. In the high CLIPScore (HC) datasets, our method shows an average improvement of +11.42\% in Acc-RN50 and +0.72\% in Acc-CLIP compared to Real Guidance (RG), and +13.87\% and +0.64\% compared to DA-Fusion~(DF). These improvements are also statistically significant, with p-values of {p\textless 0.01} when compared to baselines. In contrast, in the low CLIPScore~(LC) datasets, our method still shows an average improvement of +1.70\% in Acc-RN50 and +0.14\% in Acc-CLIP compared to RG, but a slight decrease of -0.51\% in Acc-RN50 and -1.43\% in Acc-CLIP when compared to DF. However, the improvement over RG in LC datasets remains statistically significant (p\textless 0.05), and the differences compared to DF are not statistically significant ({p\textgreater 0.1}). These results demonstrate that our method provides significant performance improvements over RG across all datasets. Additionally, while DF requires a fine-tuning stage that takes longer than generating all of our synthetic images, our approach still outperforms it in HC and shows only minimal performance differences in LC, further highlighting the practicality of our method.

\begin{table}
  \caption{Comparison of average accuracy with baselines in 1-shot setting. All values in the table represent the average of 3 trials, with the standard deviation included.}
  \label{tab:acc}
  \centering
  \resizebox{\columnwidth}{!}{ 
  \begin{tabular}{@{}ccccccc@{}}
    \toprule
    & \multicolumn{2}{c}{HC} & \multicolumn{2}{c}{LC}\\
    Methods & Acc-RN50 & Acc-CLIP & Acc-RN50 & Acc-CLIP\\
    \midrule
    RG & $0.683 \pm 0.035$ & $0.890 \pm 0.041$ & $0.504 \pm 0.019$ & $0.817 \pm 0.003$ \\
    DF & $0.658 \pm 0.050$ & $0.891 \pm 0.036$ & \textbf{0.526 \(\pm\) 0.023} & \textbf{0.833 \(\pm\) 0.021} \\
    \midrule
    {\bf Ours} & \textbf{0.797 \(\pm\) 0.023} & \textbf{0.897 \(\pm\) 0.034} & $0.521 \pm 0.014$ & $0.819 \pm 0.011$ \\
  \bottomrule
  \end{tabular}
  }
\end{table}
\clearpage

\section{Hyperparameters}
\label{sec:appendixhyper}

\begin{table}
  \caption{Hyperparameters Settings.}
  \label{tab:hyperparameter}
  \centering
  \begin{tabular}{@{}lll@{}}
    \toprule
    Hyperparameters & Values \\
    \midrule
    Synthetic Probability $p$  & 0.5 \\
    Number of Prompts $M$ & 10 \\
    GPT4 Prompt Temperature & 1.0 \\
    GPT4 Prompt Top Probability & 1.0 \\
    GPT4 Prompt Frequency Penalty & 0.0 \\
    GPT4 Prompt Presence Penalty & 0.0 \\
    Stable Diffusion Checkpoint  & CompVis/stable-diffusion-v1-4\\
    Stable Diffusion Guidance Scale & 7.5\\
    Stable Diffusion Image Size & 512\\
    IP-Adapter Checkpoint & \textnormal{ip-adapter\_sd15}\\
    Adaptive Guidance Scaling $\alpha$ & 0.3 \\
    Adaptive Guidance Scaling \(\text{min}_\text{low}\) & 0.7 \\
    Adaptive Guidance Scaling \(\text{max}_\text{low}\) & 0.9 \\
    Adaptive Guidance Scaling \(\text{min}_\text{high}\) & 0.1 \\
    Adaptive Guidance Scaling \(\text{max}_\text{high}\) & 0.4 \\
    Classifier Model & \{CLIP-ViT-B/16, ResNet50\}\\
    Classifier Optimizer & \{AdamW, Adam\}\\
    Classifier Training Epochs & 50 \\
    Classifier Learning Late & \{0.0002, 0.0001\} \\
    Classifier Image Size & 224 \\ 
    Classifier Batch Size & 32\\
  \bottomrule
  \end{tabular}
\end{table}

\clearpage

\section{LLM Prompt details}
\label{sec:appendixprompt}

\begin{figure}
    \centering
    \includegraphics[width=6.5cm]{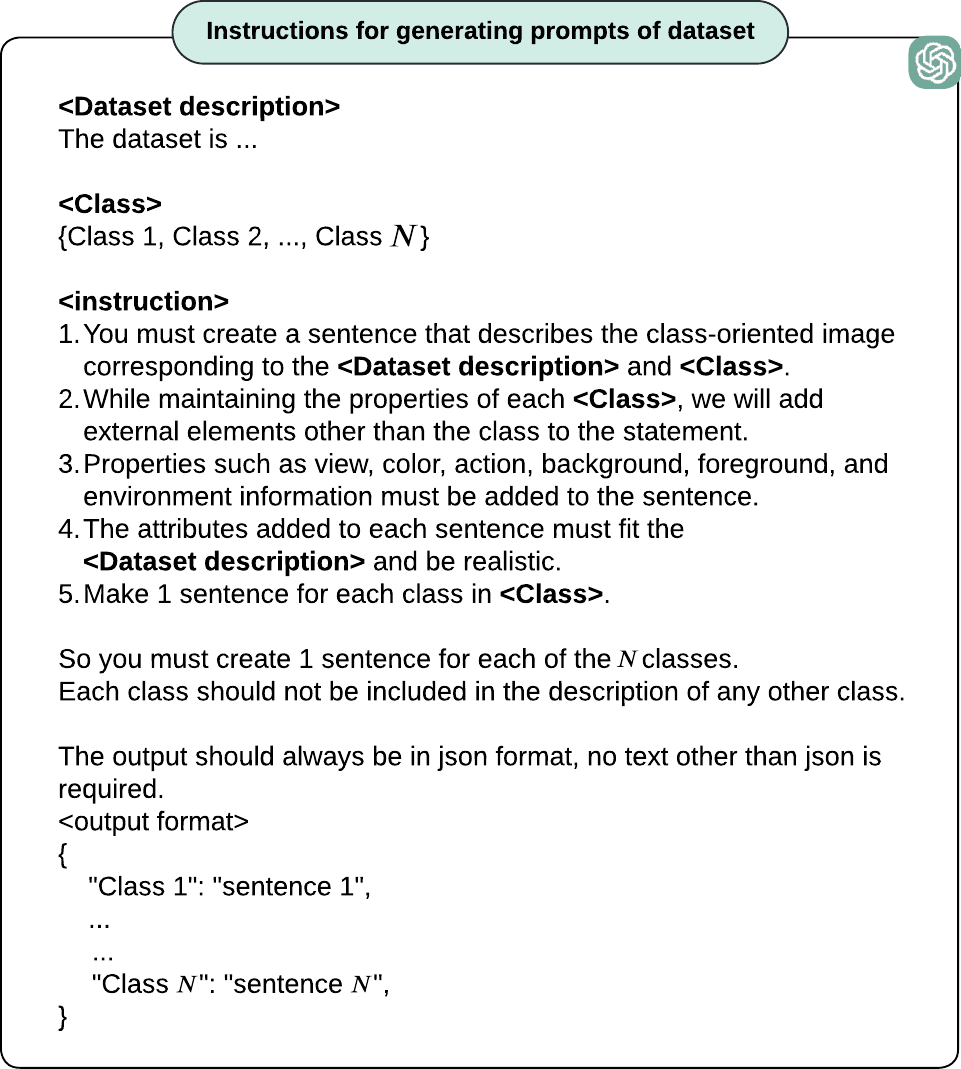}
    \caption{We use the LLM to generate one sentence for each class, iterating this process $M$ times.}
    \label{fig:prompts}
\end{figure}

\begin{table}
  \caption{Examples of dataset description. Each description was partially taken from the official websites of the respective datasets.}
  \label{tab:datasetdescription}
  \centering
  \resizebox{\textwidth}{!}{ 
  \begin{tabular}{@{}l@{\hspace{1cm}}p{14cm}@{}}
    \toprule
    Dataset & Dataset Description \\
    \midrule 
    Oxford Pets & \texttt{``We have created a 37 category pet dataset with roughly 200 images for each class. The images have a large variations in scale, pose and lighting.''}  \\
    \midrule
    Caltech-101 & \texttt{``Caltech-101 dataset is pictures of objects belonging to 101 categories. About 40 to 800 images per category. Most categories have about 50 images. The size of each image is roughly 300 x 200 pixels. We have carefully clicked outlines of each object in these pictures.''} \\
    \midrule
    Flowers102 & \texttt{``We have created a 102 category dataset, consisting of 102 flower categories. The flowers chosen to be flower commonly occuring in the United Kingdom. Each class consists of between 40 and 258 images.''} \\
  \bottomrule
  \end{tabular}
}
\end{table}

\end{document}